\documentclass{article}

\usepackage[final]{nips_2017}

\usepackage[utf8]{inputenc} % allow utf-8 input
\usepackage[T1]{fontenc}    % use 8-bit T1 fonts
\usepackage{hyperref}       % hyperlinks
\usepackage{url}            % simple URL typesetting
\usepackage{booktabs}       % professional-quality tables
\usepackage{amsfonts}       % blackboard math symbols
\usepackage{nicefrac}       % compact symbols for 1/2, etc.
\usepackage{microtype}      % microtypography
\usepackage{xcolor}      % 
\usepackage{amsmath}
\usepackage{amsthm}
\usepackage{graphicx}
\usepackage{xspace}
\usepackage{wrapfig}
\usepackage{hyperref}

\DeclareMathOperator*{\argmin}{argmin}

\newcommand{\E}{\mathbb{E}}

\newcommand{\Fb}{\F_{\beta}}

\newcommand{\bl}{\beta_l}
\newcommand{\el}{\varepsilon_l}

\newcommand{\Y}{\mathcal{Y}}

\newcommand{\F}{\mathcal{F}}
\newcommand{\M}{\mathcal{M}}

\setcitestyle{numbers,square,comma}
\newcommand{\figref}[1]{Figure~\ref{#1}\xspace}

\title{Efficient human-like semantic representations\\ via the Information Bottleneck principle}

\author{
  Noga Zaslavsky\thanks{Author for correspondence: \texttt{noga.zaslavsky@mail.huji.ac.il}\newline
  \vspace{-5pt}
  }\\
  Hebrew University\\
  UC Berkeley\\
  \And
  Charles Kemp\\
  Carnegie Mellon University\\
  \And
  Terry Regier\\
  UC Berkeley\\
  \And
  Naftali Tishby\\
  Hebrew University\\
}

\begin{document}

\maketitle

\begin{abstract}
Maintaining efficient semantic representations of the environment is a major challenge both for humans and for machines. While human languages represent useful solutions to this problem, it is not yet clear what computational principle could give rise to similar solutions in machines. In this work we propose an answer to this open question. We suggest that languages compress percepts into words by optimizing the Information Bottleneck (IB) tradeoff between the complexity and accuracy of their lexicons. We present empirical evidence that this principle may give rise to human-like semantic representations, by exploring how human languages categorize colors. We show that color naming systems across languages are near-optimal in the IB sense, and that these natural systems are similar to artificial IB color naming systems with a single tradeoff parameter controlling the cross-language variability. In addition, the IB systems evolve through a sequence of structural phase transitions, demonstrating a possible adaptation process. This work thus identifies a computational principle that characterizes human semantic systems, and that could usefully inform semantic representations in machines.
\end{abstract}

\section{Introduction}

Efficiently representing a complex environment using words is a major challenge for any cognitive system, whether biological or artificial \cite{Harnad1990,Moritz2017}. Human languages reflect different solutions to this problem, as they vary in their word meanings. Nonetheless, they all exhibit useful semantic representations and obey several universal constraints \cite{Croft2003,Berlin1969}. This suggests that there might be a general principle that gives rise to efficient semantic representations, while allowing variability along some dimensions to accommodate language-specific needs. Such a principle could advance our understanding of possible forces that may shape natural languages and could potentially be used to inform useful human-like semantic representations in machines. Here we suggest that languages compress percepts into words through the Information Bottleneck (IB) principle \cite{Tishby1999}.

IB is a general method for efficiently extracting relevant information that one variable contains about another. It was originally used to quantify and identify semantic relations between words \cite{Pereira1993,Slonim2001}, and was also suggested as a principle for learning efficient representations in biological neural networks \cite{Bialek2006,Palmer2015,Rubin2016} as well as in artificial neural networks \cite{Tishby2015,Ziv2017}. However, so far it has not been clear how to use these applications of IB to gain a better understanding of how human-like semantic representations emerge from a need to communicate about the environment. From a cognitive perspective, a need for efficient communication is emerging as a leading principle for explaining word meanings across languages \cite{Regier2015,Kemp2018}. However, this cognitive approach has not previously been cast in terms of an independently motivated computational framework that is applicable to many machine learning tasks. Here we bring these two approaches together, and formulate the IB principle as a communication game between two agents, in which word meanings are grounded in human perception.

\begin{figure}[h!]
\begin{centering} 
\includegraphics[scale=0.4,clip]{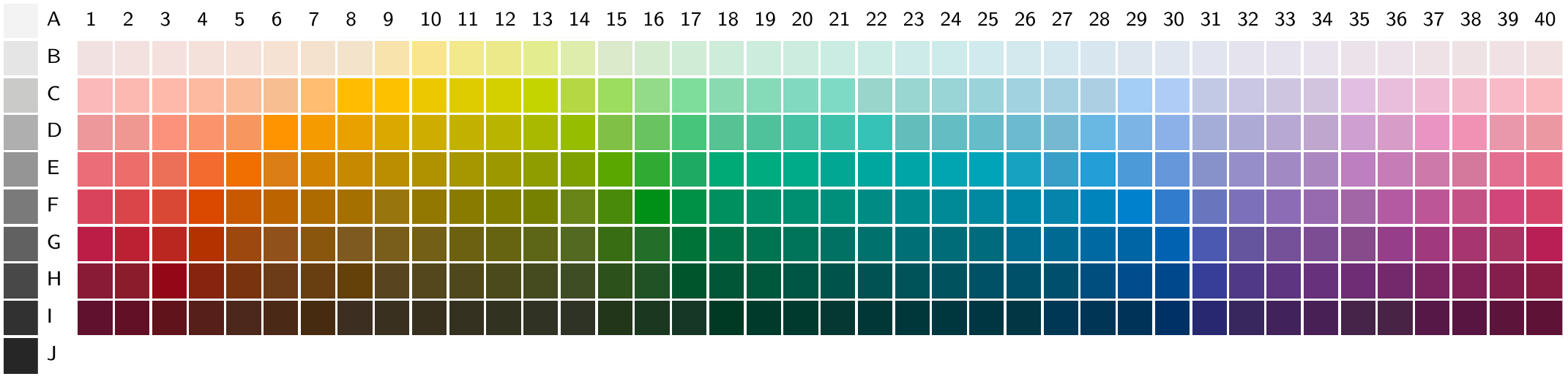}
\par\end{centering}
\caption{
\label{fig:WCS-colors}
The WCS stimulus palette.
}
\end{figure}

We present evidence that this computational principle gives rise to human-like semantic representations by studying how human languages around the world categorize colors. This is an important case study in cognitive science \cite{Berlin1969}, which also has applications in machine learning \cite{McMahan2015,Kawakami2016}. Our primary data source is the World Color Survey (WCS), which contains color naming data from 110 languages of non-industrialized societies \cite{Cook2005}. Native speakers of each language provided names for the 330 color chips shown in \figref{fig:WCS-colors}. We also analyzed color naming data from American English \cite{Lindsey2014} against the same stimulus array.

For each language $l$ we estimated a color-naming distribution $q_l(w|c)$, where $w$ is a color term and $c$ is a color chip, by the empirical distribution obtained from averaging over the responses of all participants of language $l$ (see data rows in \figref{fig:qualitative} for example).

\section{Communication model}
\label{sec:model}

\begin{figure}[h!]
\centering
\includegraphics[scale=0.25,clip]{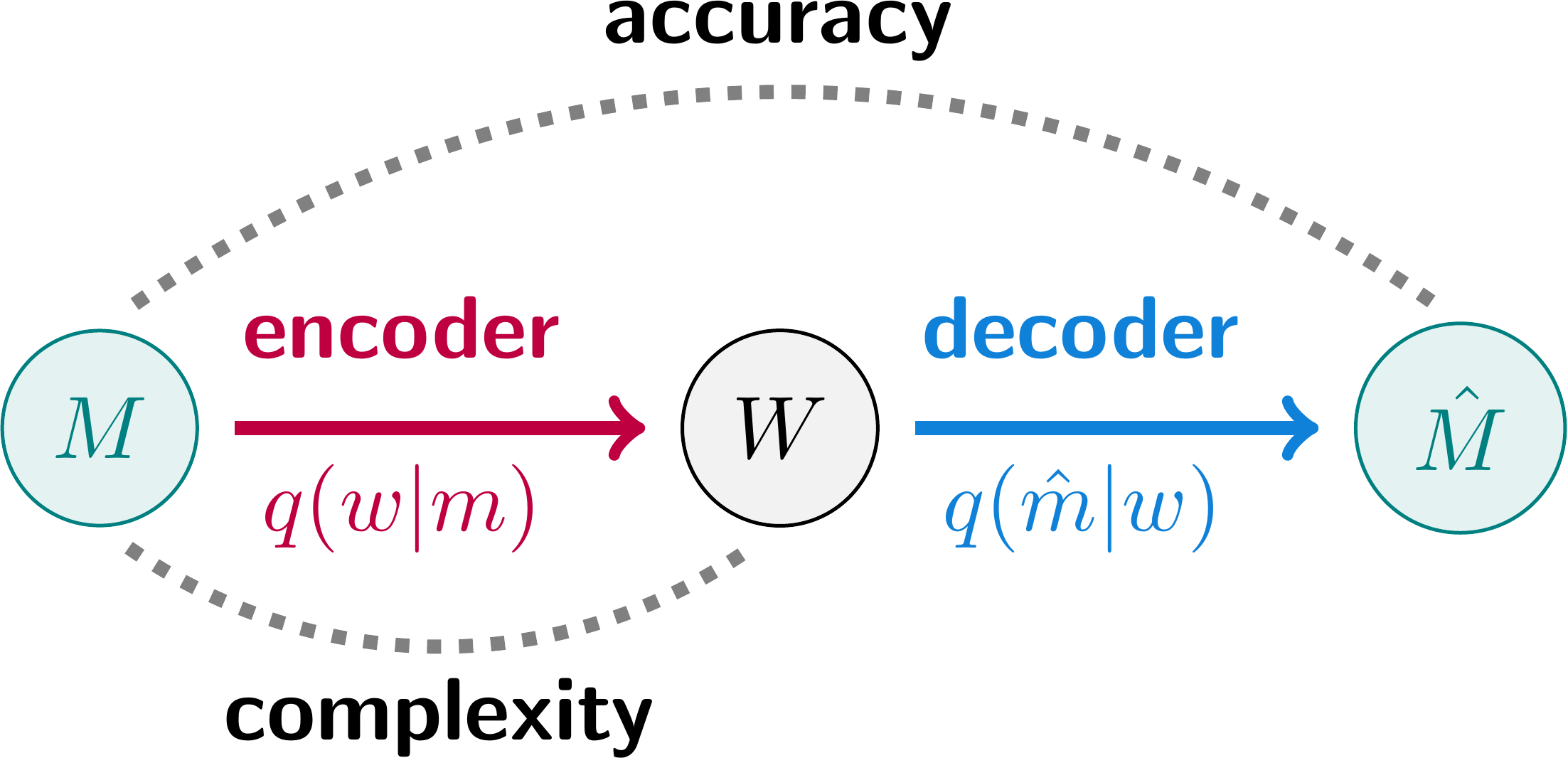}
\caption{Illustration of the semantic IB.
\label{fig:communication-model}
}
\end{figure}

We consider a communication game between a speaker and a listener, 
where the messages that the speaker wishes to communicate are distributions over the environment (\figref{fig:communication-model}). We describe the environment by a set of objects, $\Y$, that can be perceived by both parties, and define a \emph{meaning} by a distribution $m(y)$ over $\Y$. Given $m$, the speaker may think of a particular object in the environment, $Y \sim m(y)$. She would like to communicate $m$ so that the listener could think about the environment in a similar way.

We assume a \emph{cognitive source} that generates intended meanings for the speaker. This source is defined by a distribution $p(m)$ over a set of meanings, $\M$, that the speaker can represent.
The speaker communicates her intended meaning $m$ by producing a word $w$ taken from a lexicon of size $K$. We allow her to pick words according to a non-deterministic naming policy, $q(w|m)$. This policy can be seen as an \emph{encoder} because it compresses her meanings about the environment into words. The listener receives $w$ and interprets it as $\hat{m}$ based on her \emph{decoder}, $q(\hat{m}|w)$. Since this work concerns the efficiency of color naming, we assume an ideal listener that deterministically interprets $w$ as meaning $\hat{m}_w(y) = \sum_{m\in\M} m(y)q(m|w)$. Notice that $\hat{m}_w$ is the posterior distribution of $Y$ given $w$.

\paragraph{Perceptually grounded color meanings.}

To account for color naming data, we restrict the environment to the WCS palette.
We assume that each color chip corresponds to a unique meaning, $m_c$. Following a similar approach as \cite{Regier2015,Regier2007}, we ground these distributions in existing models of human color perception by representing colors in the 3-dimensional CIELAB space. We assume that each $m_c$ is an isotropic Gaussian in this space, namely
$m_{c}(y) \propto \exp\left(-\frac1{2\sigma^2}\|y-c\|^2\right)\,$. 
The scale of these Gaussians reflects the level of perceptual uncertainty. We take $\sigma$ to be a distance in which two colors can be distinguished comfortably, and determined it based on the results reported in \cite{Mokrzycki2012}.

\paragraph{Estimation of the cognitive source.}

In many cases it is not clear what process generates meanings for the speaker. A natural method for estimating the source distribution is by the least informative prior, which is closely related to reference priors in Bayesian inference~\cite{Bernardo1979,Berger2009}. For each language $l$ we evaluated the reference prior $p_l(c)$ with respect to its naming data. These priors vary across languages, and may reflect different communicative needs \cite{Gibson2017}. However, to simplify our model and reduce the number of parameters, we assume a single cognitive source that is shared among all languages. This source is defined by averaging over languages, i.e. by $p(m_c)=\frac{1}{L}\sum_l p_l(c)$ where $L$ is the number of languages.

\section{Information-theoretic bounds on semantic efficiency}

From an information-theoretic perspective, an efficient encoder minimizes its complexity by compressing the intended message $M$ as much as possible, while maximizing the accuracy of the interpretation $\hat{M}$ (\figref{fig:communication-model}). In the special case where messages are distributions, this optimization problem is captured by the Information Bottleneck (IB) principle \cite{Tishby1999,Harremoes2007}. Notice that the communication model defined in section \ref{sec:model} corresponds to the Markov chain $Y-M-W-\hat{M}$, where $Y$ reflects how the speaker thinks about the environment. The IB principle in this case is
\begin{equation}
\min_{q(w|m)} \Fb[q] : \Fb[q] = I_q(M;W) - \beta I_q(Y;W),\label{eq:comp-acc-tradeoff}
\end{equation}
where $I_q(M;W)$ corresponds to the informational complexity of the speaker's encoder, $I_q(W;Y)$ corresponds to the informativeness of the communication, and $\beta$ is the tradeoff between them. The informativeness term is directly related to the ability of the listener to accurately interpret $M$ since $ I_q(W;Y) = I(M;Y) - \E_q[D[M\|\hat{M}]]$,
where $D[\cdot\|\cdot]$ is the KL-divergence. This identity implies that maximizing $I_q(W;Y)$ w.r.t. $q(w|m)$ is equivalent to minimizing $\E_q[D[M\|\hat{M}]]$.

\begin{figure}[b!]
\centering
\includegraphics[scale=.57,clip]{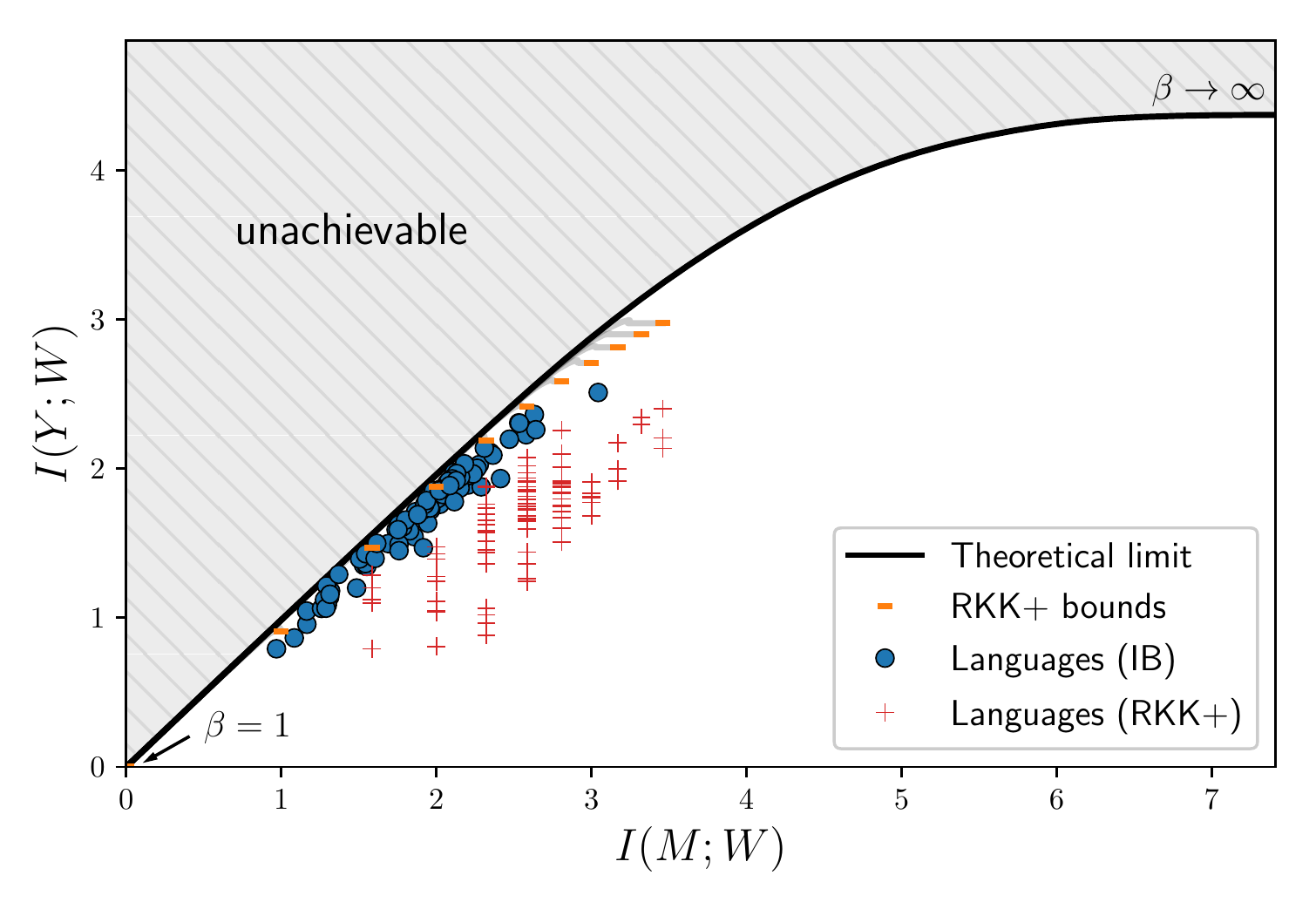}
\vspace{-10pt}
\caption{
Color naming across languages is near the information-theoretic limit (IB curve).
\label{fig:phase_space}
}
\end{figure}

Every language $l$, defined by an encoder $q_l(w|m)$, attains a certain level of complexity and a certain level of accuracy. These two quantities are plotted one against the other on the information plane shown in \figref{fig:phase_space}. The IB curve (black) is the theoretical limit defined by the set of artificial languages that optimize Eq.\eqref{eq:comp-acc-tradeoff} for different values of $\beta$.
When $\beta\rightarrow\infty$ each $m_c$ is mapped to a unique word, and when $0\le\beta\le1$ the solution of Eq.\eqref{eq:comp-acc-tradeoff} is non-informative, i.e. $I(M;W)=0$, which can be achieved by using only a single word. In between, as $\beta$ increases from 1 to $\infty$, the \emph{effective} lexicon size of the artificial IB languages changes.

\section{Results}

If human languages are shaped by a need to maintain IB efficient representations, then for each language $l$ there should be a tradeoff $\bl$ for which $l$ is close to the optimal $\F_{\bl}^*$, namely $\Delta\F_{\bl} =  \F_{\bl}[q_l] - \F_{\bl}^*$ is small. A natural way to predict $\bl$ is by $\bl=\argmin_{\beta}\Delta\Fb$. To evaluate the similarity between the artificial IB language defined by $q_{\bl}(w|m)$ and natural language defined by $q_l(w|m)$, we use a generalization of the normalized information distance \cite{Xuan2010} to soft clusterings, called gNID \cite{Zaslavsky2018}.

To control for overfitting and to challenge the ability of our approach to generalize to unseen languages, we performed 5-fold cross validation over the languages that are used for estimating the cognitive source. In addition, we consider as a baseline for comparison a similar model in which all settings are the same but efficiency is evaluated according to the principle proposed by Regier, Kemp and Kay in \cite{Regier2015}. We refer to this alternative model as RKK+. Both IB and RKK+ measure accuracy by $\E[D[M\|\hat{M}]]$, however in RKK+ complexity is measured by the number of frequent color terms (here we take the log). In addition, RKK+ evaluates each language w.r.t. to its optimal solution at the same complexity. We therefore consider also a variant of our IB approach, which we call C-IB, in which $\bl$ is estimated such that the IB complexity measure is constrained in the same way. That is, in C-IB it holds that $I(M;W)$ is the same for $q_l$ and for $q_{\bl}$. We evaluate the deviation from optimality for all three models by $\el=\frac1\bl\Delta\F_{\bl}$, where in RKK+ and C-IB this measure is reduced to the difference in accuracy regardless of $\bl$.

Table \ref{tab:kf_cv} shows the results of the 5-fold cross validation. IB and C-IB achieve very similar scores, although gNID is slightly better for IB. In addition, C-IB achieves $74\%$ improvement in $\el$ and $55\%$ improvement in gNID compared to RKK+. Similar results are obtained when the cognitive source is estimated from all folds. Therefore, the IB curve and RKK+ bounds shown in \figref{fig:phase_space} are evaluated for the source distribution estimated from the full data.

\begin{table}[h!]
\centering
\begin{tabular}{lcc}
Principle 		& $\el$ 			& gNID			\\
\midrule
IB 			& 0.18 ($\pm$0.07) 	& 0.18 ($\pm$0.10)	\\
C-IB 			& 0.18 ($\pm$0.07) 	& 0.21 ($\pm$0.08)	\\
RKK+ 		& 0.70   ($\pm$0.23) & 0.47 ($\pm$0.10) 	\\
\bottomrule
\end{tabular}
\vspace{10pt}
\caption{
Numbers correspond to averages over left-out languages $\pm$1 SD. Lower values are better.
\label{tab:kf_cv}
}
\vspace{-4pt}
\end{table}

\begin{figure}[t!]
\centering
\includegraphics[scale=.72,trim={2mm 7mm 0 0},clip]{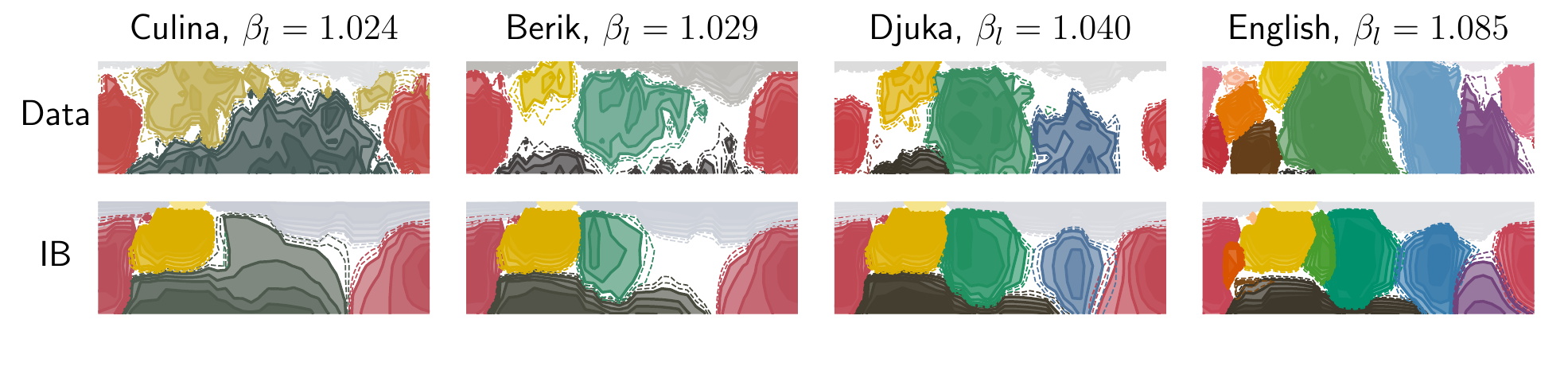}
\caption{  
\label{fig:qualitative}
Similarity between four languages (data rows) for example and their corresponding artificial IB languages at $\bl$ (IB rows). 
Each plot shows the contours of the naming distribution with level sets 0.5-0.9 (solid lines) and 0.4-0.45 (dashed lines). Colors correspond to the color-centroids of each category.
\vspace{-5pt}
}
\end{figure}

\figref{fig:phase_space} and the small $\el$ score for IB show that the efficiency of color naming in all languages is near the information-theoretic limit. In addition, IB's low gNID score suggests that natural color naming systems are similar to the artificial IB color naming systems. This is also supported by a visual inspection of the data (\figref{fig:qualitative}).

We wish to emphasize that the qualitatively different solutions along the IB rows in \figref{fig:qualitative} are caused solely by the small changes in $\beta$. This single parameter controls the complexity, accuracy and effective lexicon size of the IB encoders.
The IB categories evolve through a sequence of structural phase transitions as $\beta$ increases, in which the number of distinguishable color categories changes. This process is similar to the deterministic annealing procedure for clustering \cite{Rose1990,Rose1998}.
This demonstrates a process in which the artificial IB languages may adapt to changing conditions, that also resembles cognitive theories of language evolution~\cite[e.g.][]{Berlin1969,Levinson2000}.
 
 \section*{Summary}
 
We have shown that a need to maintain information-theoretically efficient semantic representations can account for how natural languages represent colors; the same principle could also be used to inform human-like semantic representations of color in machines. The generality of our methods suggests that this approach may also be applied to other perceptually-grounded semantic domains. The only component in our framework that is specific to color is the meaning space.

\subsubsection*{Acknowledgments}
We thank Delwin Lindsey and Angela Brown for kindly sharing their English color-naming data with us.
This study was supported by the Gatsby Charitable Foundation. N.Z. was supported by the IBM Ph.D. Fellowship Award.

\bibliography{bibliography}

\end{document}